\documentclass{article}

\usepackage{graphicx} 
\usepackage{subfigure} 

\usepackage{natbib}

\usepackage{algorithm}
\usepackage{algorithmic}

\usepackage{hyperref}

\usepackage{noticml} 

\usepackage{epsfig}

\usepackage{amsmath}
\usepackage{nicefrac}
\usepackage{amsfonts}

\def\beq#1\eeq{\begin{equation}#1\end{equation}}
\def\beqn#1\eeqn{\begin{displaymath}#1\end{displaymath}}
\def\beqa#1\eeqa{\begin{eqnarray}#1\end{eqnarray}}

\newcommand{\diag}{\textrm{diag}}

\newcommand{\X}{{\cal{X}}}

\renewcommand{\citet}[1]{\aunpcite{#1}~\yrcite{#1}} 

\newcommand{\tO}{\tilde{O}}
\newcommand{\R}{\mathbb{R}}
\newcommand{\NN}{\mathcal{N}}
\newcommand{\Ep}[2]{{{\mbox{\bf E}}_{#1}\!\left[{#2}\right]}}

\newcommand{\norm}[1]{\left\lVert{#1}\right\rVert}

\newcommand{\rank}{\textit{rank}}
\newcommand{\tc}{\textit{tc}}
\newcommand{\dist}{{\cal{D}}} 
\newcommand{\trnorm}[1]{\norm{#1}_{\text{tr}}}
\newcommand{\trpqnorm}[2]{\norm{#1}_{\text{tr}\left(#2\right)}}
\newcommand{\frnorm}[1]{\norm{#1}_{\text{F}}}

\icmltitlerunning{Learning with the Weighted Trace Norm}

\begin{document} 

\twocolumn[
\icmltitle{Collaborative Filtering in a Non-Uniform World: \\ Learning
  with the Weighted Trace Norm}

\icmlauthor{Ruslan Salakhutdinov}{rsalakhu@mit.edu}
\icmladdress{Brain and Cognitive Sciences and CSAIL, Massachusetts Institute of Technology}
\icmlauthor{Nathan Srebro}{nati@ttic.edu}
\icmladdress{Toyota Technological Institute--Chicago}


\vskip 0.3in
]

\begin{abstract} 
  We show that matrix completion with trace-norm regularization can be
  significantly hurt when entries of the matrix are sampled
  non-uniformly.  We introduce a weighted version of the trace-norm
  regularizer that works well also with non-uniform sampling.  Our
  experimental results demonstrate that the weighted trace-norm
  regularization indeed yields significant gains on the (highly
  non-uniformly sampled) Netflix dataset.
\end{abstract} 

\section{Introduction}
Trace-norm regularization is a popular approach for matrix completion
and collaborative filtering, motivated both as a convex surrogate to
the rank \cite{fazel,candes2009power} and in terms of a regularized
infinite factor model with connections to large-margin
norm-regularized 
learning \cite{mmmf,journals/corr/abs-0710-2848,Abernethy2009,SalMnih08}.

Current theoretical guarantees on using the trace-norm for matrix
completion all assume a uniform sampling distribution over entries of
the matrix
\cite{srebro2005rank,candes2009power,candes-exact,candes2009power,recht-preprint}.
In a collaborative filtering setting, where rows of the matrix
represent e.g.~users and columns represent e.g.~movies, this
corresponds to assuming all users are equally likely to rate movies
and all movies are equally likely to be rated.  This of course cannot
be further from the truth, as in any actual collaborative filtering
application, some users are much more active then others and some
movies are rated by many people while others are much less likely to
be rated.

In Section \ref{sec:fails} we show, both analytically and through 
simulations, that this is not a deficiency of the proof techniques
used to establish the above guarantees. Indeed, a non-uniform sampling
distribution can lead to a significant deterioration in prediction 
quality and an increase in the sample complexity.  Under non-uniform
sampling, as many as $\Omega(n^{4/3})$ samples might be needed for
learning even a simple (e.g. orthogonal low rank) $n \times n$
matrix.  This is in sharp contrast to the uniform sampling case, in which
$\tO(n)$ samples are enough.  It is important to note that if the rank
could be minimized directly, which is in general not computationally
tractable, $\tO(n)$ samples would be enough to learn a low-rank model
even under an arbitrary non-uniform distribution.

In Section \ref{sec:weighted} we suggest a correction to the
trace-norm regularizer, which we call the {\em weighted} trace-norm,
that takes into account the sampling distribution.  This correction is
motivated by our analytic analysis and we discuss how it corrects the
problems that the unweighted trace-norm has with non-uniform sampling.  We
then show how the weighted trace-norm indeed yields a significant
improvement on the (highly non-uniformly sampled)  Netflix dataset.

\section{Complexity Control in terms of Matrix
  Factorizations}\label{sec:mf}
Consider the problem of predicting the entries of some
unknown target matrix $Y \in \R^{n \times m}$ based on a random subset $S$ of
observed entries $Y_S$. 
For example, $n$ and $m$ may 
represent the number of users and 
the number of movies, and $Y$ may represent a matrix of 
partially observed rating values.  
Predicting elements of $Y$  can be done  by finding a matrix $X$
minimizing the training error, here measured as a squared error, and
some measure $c(X)$ of complexity.  That is, minimizing either: 
\begin{equation}\label{eq:clambda}
  \min_{X} \frnorm{X_S-Y_S}^2 + \lambda c(X)
\end{equation}
or:
\begin{equation}\label{eq:cC}
  \min_{c(X) \leq C} \frnorm{X_S-Y_S}^2, 
\end{equation}
where $Y_S$, and similarly $X_S$, denotes 
the matrix ``masked'' by $S$:
\begin{equation}
  \label{eq:YS}
  (Y_S)_{i,j} =
  \begin{cases}
    Y_{i,j} & \text{if $(i,j) \in S$} \\
0 & \text{otherwise}.
  \end{cases}
\end{equation}
For now we ignore possible repeated entries in $S$.
We will also assume that $n \leq m$ without loss of generality.

The two formulations \eqref{eq:clambda} and \eqref{eq:cC} are
equivalent up to some (unknown) correspondence between $\lambda$ and
$C$, and we will be referring to them interchangeably at our
convenience.

\subsection{Low Rank Factorization}
A basic measure of complexity is the rank of $X$, 
corresponding to the
minimal dimensionality $k$ such that $X = U^{\top}V$ for some 
$U \in \R^{k\times n}$ and $V \in \R^{k \times m}$. 

Directly constraining the rank of $X$ forms one of the most popular
approaches to collaborative filtering.  Training such a model amounts
to finding the best rank-$k$ approximation to the observed target
matrix $Y$ under the given loss function.  However, the rank is
non-convex and hard to minimize. It is also not clear if a strict
dimensionality constraint is most appropriate for measuring the
complexity.

\subsection{Trace-norm Regularization}
Lately, methods regularizing the {\em norm} of the factorization
$U^{\top}V$, rather then its dimensionality, have been advocated and were
shown to enjoy considerable empirical success
\cite{rennie2005fast,SalMnih08}.  This is captured by measuring complexity
in terms of the {\em trace-norm} of $X$, which can be defined
equivalently either as the sum of the singular values of $X$, or as
\cite{fazel}:
\begin{equation}
  \label{eq:trnorm}
  \trnorm{X} = \min_{X=U'V} \frac{1}{2}(\frnorm{U}^2+\frnorm{V}^2).
\end{equation}
Note that the dimensionality of $U$ and $V$ in \eqref{eq:trnorm} is not
constrained.  Beyond the modeling appeal of norm-based, rather then
dimension-based, regularization, the trace-norm is a convex function
of $X$ and so can be minimized by either local search or more
sophisticated convex optimization techniques.

\subsection{Scaling of the Trace-norm} 
It will be useful for us to consider the scaling of the trace-norm
with the size of the matrix $X$. This will allow us, for example, to
understand the magnitude of the bound $C$ we can expect to put on the
trace-norm in the formulation \eqref{eq:cC}.

The rank, as a measure of complexity, does not scale with the size of
the matrix.  That is, even very large matrices can have low rank.
Viewing the rank as a complexity measure corresponding to the number of
underlying factors, if data is explained by e.g.~two factors, then no
matter how many rows (``users'') and columns (``movies'') we consider,
the data will still have rank two.

The trace-norm, however, does inherently scale with the size of the
matrix.  To see this, note that the trace-norm is the $\ell_1$ norm of
the spectrum, while the Frobenius norm is the $\ell_2$ norm of the
spectrum, yielding:
\begin{equation}
  \label{eq:FTRF}
  \frnorm{X} \leq \trnorm{X} \leq \frnorm{X} \sqrt{\rank(X)} \leq
  n\frnorm{X}, 
\end{equation}
where in the second inequality we used the fact that the number of
non-zero singular values is equal to the rank. The Frobenius norm
certainly increases with the size of the matrix, since the magnitude
of each element does not decrease when we have more elements, and so
the trace-norm will also increase. The above suggests measuring
the trace-norm relative to the Frobenius norm.  Without loss of generality,
consider each target entry to be of roughly unit 
magnitude\footnote{Any other constant magnitude will only result in some constant
scaling},  e.g.~$\pm1$,
and so in order to fit $Y$ each entry of $X$ must also be of
roughly unit magnitude. This suggests scaling the trace-norm by
$\sqrt{nm}$.  More specifically, we study the trace-norm
through the complexity measure: 
\begin{equation}
  \label{eq:tc}
  \tc(X) = \frac{\trnorm{X}^2}{nm}, 
\end{equation}
which puts the trace-norm on a comparable scale to the rank.  In
particular, when each entry of $X$ is, on-average, of unit magnitude
(i.e.~has unit variance),  in which case  $\frnorm{X}=\sqrt{nm}$, we have:
\beqa
1 \leq \tc(X) \leq \rank(X) \leq n.
\eeqa  

To further understand the trace-norm complexity control, consider 
``orthogonal'' low-rank matrices $U~\kern-2pt\in~\kern-2pt\R^{k \times n}$ 
and $V~\kern-2pt\in~\kern-2pt\R^{k \times m}$, such that $Y=U^{\top}V$ and where the entries of $U$ and $V$ 
are i.i.d.~$\NN(0,1/\sqrt{k})$\footnote{The important issue here is the
  orthogonality and the norm uniformity, not the randomness. But we
  find it easier to think of the orthogonality in terms of an
  i.i.d.~random model.}. The matrix $Y$ is then of rank $k$, with each
entry having zero mean and unit variance (magnitude). Its 
Frobenius norm is tightly concentrated at $\frnorm{Y}=\sqrt{nm}$.
Since rows of $U$ and $V$ are orthogonal, this is essentially the
singular value decomposition, with
all $k$ singular values being equal to $\sqrt{nm/k}$.  We thus have
$\tc(X)=k$.  And so at least in the orthogonal case,
$\tc(X)=\rank(X)$.

Another place where we can see that $\tc(X)$ plays a similar role to
$\rank(X)$ is in the generalization and sample complexity guarantees
that can be obtained for low-rank and low-trace-norm learning.  Such
learning guarantees were mostly discussed in the context of Lipschitz
continuous loss functions (i.e.~functions with a bounded first
derivative), rather then the squared loss.  The squared loss has a
bounded second derivative rather then bounded first derivative and so
requires somewhat different technical tools.  Nevertheless, the main
thrust of the results is still valid.  

For Lipschitz continuous loss functions, if there is a low-rank matrix
$X^*$ achieving low average error relative to $Y$ (e.g.~if
$Y=X^*+\text{noise}$), then by minimizing the training error subject
to a rank constraint (a computationally intractable task),
$|S|=\tO(\rank(X^*)(n+m))$ samples are enough in order to guarantee
learning a matrix $X$ whose overall average error is close to that of
$X^*$ \cite{genlowrank}.  Similarly, if there is a low-trace-norm
matrix $X^*$ achieving low average error, then minimizing the training
error and the trace-norm (a convex optimization problem),
$|S|=\tO(\tc(X^*)(n+m))$ samples are enough in order to guarantee
learning a matrix $X$ whose overall average error is close to that of
$X^*$ \cite{srebro2005rank}.  In these bounds $\tc(X)$ plays precisely
the same role as the rank, up to logarithmic factors.

Without getting into the technical tools required to rigorously
establish the above sample complexity guarantees, it is useful to
understand them at a more abstract level.  In order to understand the
guarantees for low-rank learning, it is enough to consider the number
of parameters in the rank-$k$ factorization $X=U^{\top}V$.  It is easy to
see that the number of parameters in the factorization is roughly
$k(m+n)$ (perhaps a bit less due to rotational invariants). And so we
would expect to be able to learn $X$ when we have roughly this many
samples, as is indeed confirmed by the rigorous sample complexity
bounds.

For low-trace-norm learning, consider a sample $S$ of size $|S| \leq C
n$, for some constant $C$.  Taking entries of $Y$ to be of unit magnitude, 
we have $\frnorm{Y_S} = \sqrt{|S|} = \sqrt{C n}$ (Recall that $Y_S$ is
defined to be zero outside $S$). From \eqref{eq:FTRF} we therefore
have: $\trnorm{Y_S} \leq \sqrt{C n} \cdot \sqrt{n} = \sqrt{C} n$ and
so $\tc(Y_S) \leq C$.  That is, we can ``shatter'' any sample of size
$|S| \leq C n$ with $\tc(X)=C$: no matter what the underlying matrix
$Y$ is, we can always perfectly fit the training data with a low
trace-norm matrix $X$ s.t.~$\tc(X)\leq C$, without generalizing at all
outside $S$.  On the other hand, we must allow matrices with
$\tc(X)=\tc(X^*)$, otherwise we can't hope to find $\X^*$, and so we
can only constrain $\tc(X)\leq C=\tc(X^*)$.  We therefore cannot
expect to learn with less then $n\tc(X^*)$ samples.  It turns out that
this is essentially the largest random sample that can be shattered
with $tc(X)\leq C=\tc(X^*)$, and that if we have more then this many
samples we can start learning.  For our purposes here, we will mostly
just make use of non-learnability arguments of this form: if we can
shatter a random sample of size $|S|$ with a matrix $X$ have the same
complexity (e.g.~trace-norm) as our target matrix $X^*$, we cannot
hope to learn without a larger sample.

\section{Trace-Norm Under a Non-Uniform Distribution}\label{sec:fails}
In this section, we will analyze trace-norm regularized learning when
the sampling distribution is not uniform.  That is, when there is
some, known or unknown, non-uniform distribution $\dist$ over entries
of the matrix $Y$ (i.e. over index pairs $(i,j)$) and our sample $S$
is sampled i.i.d.~from $\dist$.  Of course, if $\dist$ concentrates on
only a small subset of the matrix, we have no hope of recovering rows
and columns of $Y$ on which we have zero probability of seeing an
observation.  Instead, our objective here, as is typically the case in
learning under an arbitrary distribution, is to get low average error
with respect to the same distribution $\dist$.  That is, we measure
generalization performance in terms of the weighted sum-squared-error:
\beqa \label{eq:frDD}
\norm{X-Y}^2_{\dist} &=& \Ep{(i,j)\sim\dist}{(X_{ij}-Y_{ij})^2} \nonumber \\
&=& \sum_{ij} \dist(i,j) (X_{ij}-Y_{ij})^2. \eeqa 
We first point out
that when using the rank for complexity control, i.e.~when minimizing
the training error subject to a low-rank constraint, non-uniformity
does {\em not} pose a problem.  The same generalization and learning
guarantees that can be obtained in the uniform case, also hold under
an arbitrary distribution $\dist$.  In particular,
if there is some low-rank $X^*$ such that $\norm{X^*-Y}^2_{\dist}$ is
small, then $\tO(\rank(X^*)(n+m))$ samples are enough in order to
learn (by minimizing training error subject to a rank constraint) a
matrix $X$ with $\norm{X-Y}^2_{\dist}$ almost as small as
$\norm{X^*-Y}^2_{\dist}$ \cite{genlowrank}\footnote{Actually,
  this is shown only for Lipschitz continuous loss functions, and not
  for the squared-loss, but at the very least this holds if $X$ is
  appropriately clipped.  Since formal guarantees are not the focus of
  this paper, we rather view this statement only as an indicative
  statement without stating it rigorously.}.

\begin{figure}[t]
\begin{center}
\begin{picture}(80,80)(0,0)
\put(0,0){\framebox(80,80){}}
\put(0,70){\framebox(10,10){$A$}}
\put(10,30){\framebox(40,40){$B$}}
\end{picture}
\vspace{-0.1in} 
\caption{ \label{fig:AB} The two submatrices $A$ of size
  $n_A=n^{\alpha}$ and $B$ of size $n_B=n/2$.
 }
\end{center}
\vspace{-0.2in} 
\end{figure}
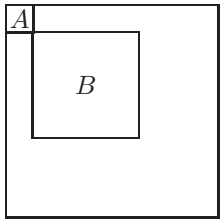

However, the same does not hold when learning using the trace-norm.
To see this, consider an orthogonal rank-$k$ square $n \times n$
matrix, and a sampling distribution which is uniform over an $n_A
\times n_A$ sub-matrix $A$, with $n_A = n^a$ (see Fig.
\ref{fig:AB}).  That is, the row (e.g.~``user'') is selected uniformly
among the first $n_A$ rows, and the column (e.g.~``movie'') is
selected uniformly among the first $n_A$ columns.  We will use $A$ to
denote the subset of entries in the submatrix, i.e.~$A = \{ (i,j) | 1
\leq i,j \leq n_A \}$, rather then the matrix itself, and so we can
say that $\dist$ is uniform on $A$.  For any sample $S$, we have: 
\begin{align}
\tc(Y_S) &= \frac{\trnorm{Y_S}^2}{n^2} \leq \frac{\frnorm{Y_S}^2
  \rank(Y_S)}{n^2} \notag\\
& \leq \frac{|S| n^a}{n^2} = \frac{|S|}{n^{2-a}}, \label{eq:tc1}
\end{align}
where we again take the entries in $Y$ to be of unit magnitude.  In
the second inequality above we use the fact that $Y_S$ is zero outside
of $A$, and so we can bound the rank of $Y_S$ by the dimensionality
$n_A = n^a$ of $A$.

Setting $a<1$, we see that we can shatter any sample of
size\footnote{Recall that $f(n)=\tilde{\omega}(g(n))$ is the same as
  $g(n)=\tilde{o}(f(n))$ and means that for all $p$ we have
  $\frac{g(n)\log^p g(n)}{f(n)}\rightarrow 0$.} $k
n^{2-a}=\tilde{\omega}(n)$ with a matrix $X$ for which $\tc(X)\kern-5pt~<~\kern-5ptk$.
When $a \leq 1/2$, the total number of entries in $A$ is less then
$n$, and so $\tO(n)$ observations are enough in order to memorize
$Y_A$.  But when $1/2<a<1$, with $\tO(n)$ observations, restricting to
even $\tc(X)<1$, we can neither learn $Y$, since we can shatter $Y_S$,
nor memorize it.  For example, when $a=2/3$ and so $n_A = n^{2/3}$, we
need roughly $n^{4/3}$ to start learning by constraining $\tc(X)$ to a
constant --- the same as we would need in order to memorize $Y_A$.
This is a factor of $n^{1/3}$ greater then the sample size needed to
learn a matrix with constant $\tc(X)$ in the uniform case.

The above arguments establish that restricting the complexity to
$\tc(X)<k$ might not lead to generalization with $\tO(k n)$ samples in
the non-uniform case.  But does this mean that we cannot learn 
a rank-$k$ matrix by minimizing the trace-norm using $\tO(k n)$
samples when the sampling distribution is concentrated on a small
submatrix?  Of course this is not the case.  Since the samples are
uniform on a small submatrix, we can just think of the submatrix A as
our entire space. The target matrix still has low rank, 
even when restricted to A, and we are back in the uniform sampling
scenario.  The only issue here is 
that $\tc(X) \leq k$, i.e. $\trnorm{X} \leq n \sqrt{k}$, is the right constraint
in the uniform observation scenario. When samples are concentrated in
$n_A$, we actually need to restrict to a much smaller trace norm,
$\trnorm{X} \leq n^a \sqrt{k}$, which will allow learning with $\tO(k
n^a)$ samples.

It is, however, easy to modify the above example and construct a
sampling distribution under which $\Omega(n^{4/3})$ samples are
required in order to learn even an ``orthogonal'' low-rank matrix, no
matter what constraint is placed on the trace-norm.  This is a
significantly large sample complexity then $\tO(k n)$, which is what
we would expect, and what is required for learning by constraining the
rank directly.

To do so, consider another submatrix $B$ of size $n_B \times n_B$ with
$n_B = n/2$, such that the rows and columns of $A$ and of $B$ do not
overlap (Fig. \ref{fig:AB}).  Now, consider a sampling distribution
$\dist$ which is uniform over $A$ with probability half, and uniform
over $B$ with probability half.  Consider fitting a noisy matrix
$Y=X^*+\text{noise}$ where $X^*$ is ``orthogonal'' rank-$k$.  In order
to fit on $B$, we need to allow a trace-norm of at least
$\trnorm{X^*_B} = \frac{n}{2} \sqrt{k}$, i.e.~allow $\tc(X)=k/4$.  But
as discussed above, with such a generous constraint on the trace-norm,
we will be able to shatter $S \subset A$ whenever $|S \cap A| = |S|/2
\leq k/4 n^{2-a}$.  Since there is no overlap in rows and columns, and
so values in the sub-matrices $A$ and $B$ are independent, shattering
$S \cap A$ means we cannot hope to learn in $A$.  Setting $a=2/3$ as
before, it seems that with $o(n^{4/3})$ samples, we cannot learn in
both $A$ and $B$: either we constrain to a trace-norm which is too
low to fit $X^*_B$ (we under-fit on $B$), or we allow a trace-norm
which is high enough to overfit $Y_{S \cap A}$.  Either way, we will
make errors on at least half the mass of $\dist$.\footnote{To make the above
  argument more precise, we should note that if we do allow high
  enough trace-norm to fit $B$, and $|S| = o(n^{4/3})$, then the
  ``cost'' of overfitting $Y_{S \cap A}$ is negligible compared to the
  cost of fitting $X^*_B$.  For large enough $n$, we would be tempted
  to very slightly deteriorate the fit of $X^*_B$ in order to ``free
  up'' enough trace-norm and completely overfit $Y_{S \cap A}$.}

\begin{figure*}[t]
\vspace{0.00in}
\hbox{ \centering \hspace{0.3in}
\setlength{\epsfxsize}{2.6in}
\epsfbox{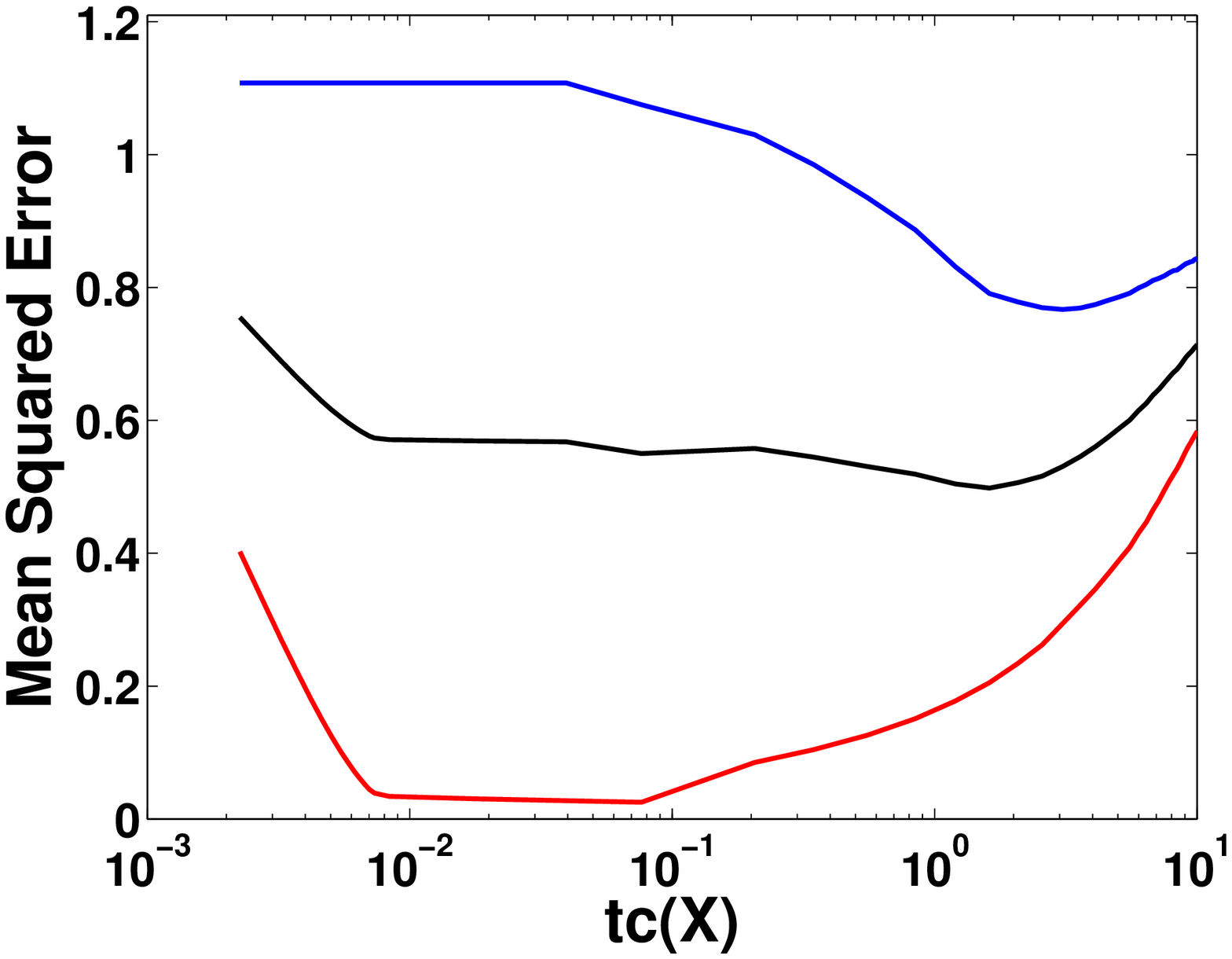}
 \hspace{0.1in}
\setlength{\epsfxsize}{2.6in}
\epsfbox{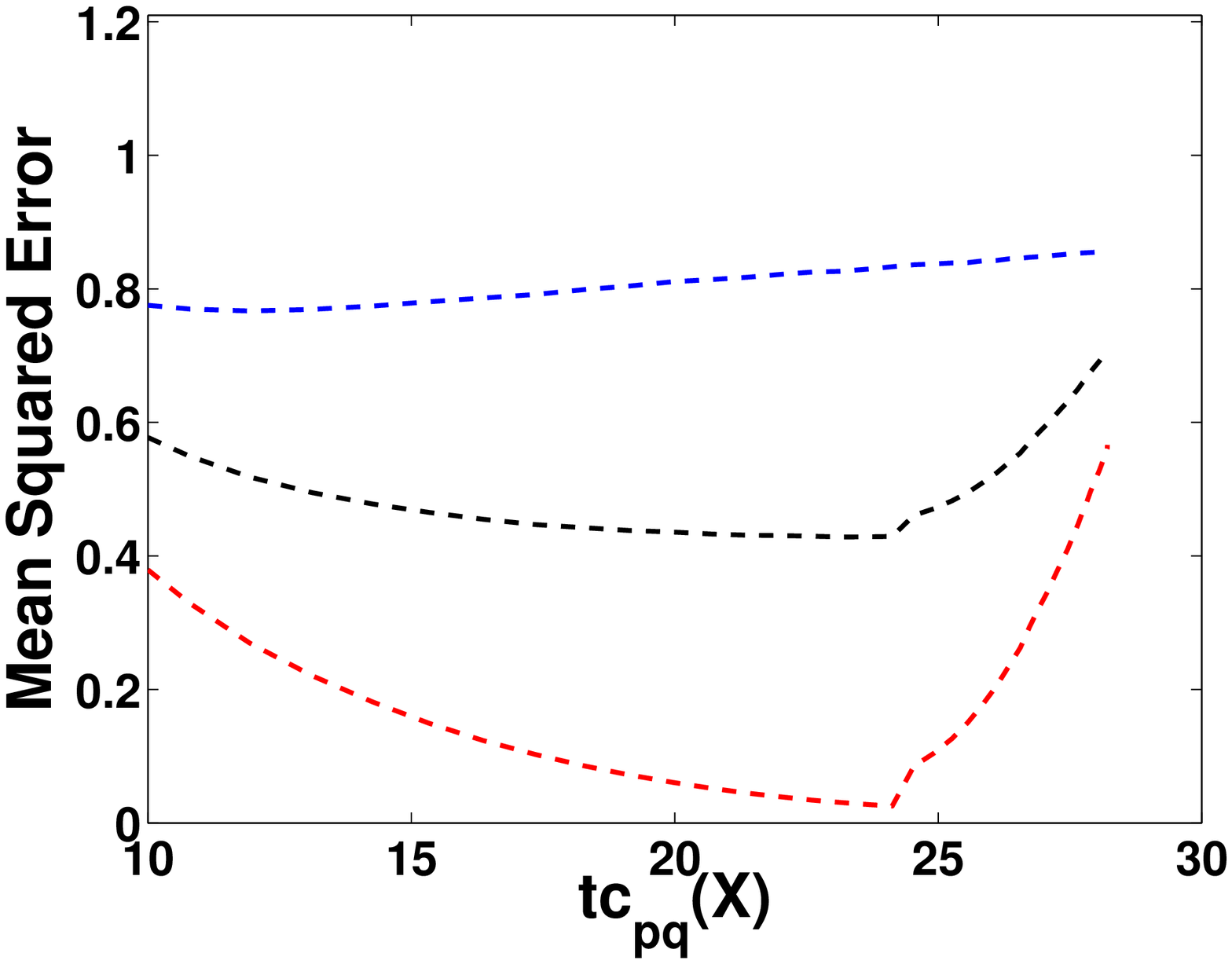}
\begin{picture}(0,0)(0,0)
\put(-155, +102){\textrm{{\bf \normalsize B}}}
\put(-155, +75){\textrm{{\bf \normalsize A+B}}}
\put(-137, +45){\textrm{{\bf \normalsize A}}}
\put(-280, +125){\textrm{{\bf \normalsize B}}}
\put(-320, +83){\textrm{{\bf \normalsize A+B}}}
\put(-280, +37){\textrm{{\bf \normalsize A}}}
\end{picture}
}
\vspace{-0.15in} 
\caption{ \label{fig:tc} \small
Mean squared error (MSE) of the learned model  
as a function of the constraint on 
$\tc(X)$ (left) and $\tc_{pq}(X)$ (right).
The black (middle) curve is the overall
MSE error, the red (bottom) curve measures only the contribution
from $A$, and the blue (top) curve measures only the contribution from
$B$. 
\vspace{-0.1in} 
}
\end{figure*}

Figure \ref{fig:tc}, left panel, precisely illustrates this phenomenon on a
simulation experiment.  For this synthetic example, we used $n_A=300$
and $n_B=4700$, with an orthogonal rank-2 matrix $X^*$ and
$Y=X^*+{\cal{N}}(0,1)$ (in case of repeated entries, the
noise is independent for each appearance in the sample).  The training
sample size was also set to $|S|$=140,000.

The three curves of Fig. \ref{fig:tc} measure the excess (test) error
$\norm{X-X^*}_{\dist}^2 = \norm{X-Y}_{\dist}^2 -
\norm{Y-X^*}_{\dist}^2$ of the learned model, as well as the error
contribution from $A$ and from $B$, as a function of the constraint on
$\tc(X)$, for the sampling distribution discussed above and a specific
sample size.
As can be seen, although it is possible to constrain $\tc(X)$ so
as to achieve squared-error of less then $0.8$ on $B$, this constraint
is too lax for $A$ and allows for over-fitting.  Constraining $\tc(X)$
so as to avoid overfitting $A$ (achieving almost zero excess test
error), leads to a suboptimal fit on $B$.

Until now we discussed learning by constraining the trace-norm, i.e. using
the formulation \eqref{eq:cC}.   It is also insightful to 
consider the
penalty view \eqref{eq:clambda}, i.e.~learning by minimizing
\begin{equation}\label{eq:trlamba1}
  \min_X \frnorm{Y_S-X_S}^2 + \lambda \trnorm{X}.
\end{equation}
First observe that the characterization \eqref{eq:trnorm} allows us to
decompose $\trnorm{X} = \trnorm{X_A} + \trnorm{X_B}$, where
w.l.o.g.~we take all columns of $U$ and $V$ outside $A$ and $B$ to be
zero.  Since we also have $\frnorm{Y_S-X_S}^2 = \frnorm{Y_{A\cap
    S}-X_{A\cap S}}^2 + \frnorm{Y_{B\cap S}-X_{B \cap S}}^2$, we can
decompose the training objective \eqref{eq:trlamba1} as:
\begin{align}
& \frnorm{Y_S-X_S}^2 + \lambda \trnorm{X} \notag\\
& = (\frnorm{Y_{A\cap S}-X_{A\cap S}}^2 + \lambda \trnorm{X_A})
\notag\\
&\quad +
  (\frnorm{Y_{B\cap S}-X_{B \cap S}}^2 + \lambda \trnorm{X_B})\notag \\
& = \left(\frnorm{Y_{A\cap S}-X_{A\cap S}}^2 + \lambda n_A
\sqrt{\tc_A(X_A)}\right) \notag\\
&\quad+
  \left(\frnorm{Y_{B\cap S}-X_{B \cap S}}^2 + \lambda n_B \sqrt{\tc_B(X_B)}\right), 
  \label{eq:trdecomp1}
\end{align}
where $\tc_A(X_A)=\trnorm{X_A}^2/n_A^2$ (and similarly $\tc_B(X_B)$)
refers to the complexity measure $\tc(\cdot)$ measured relative to the
size of $A$ (similarly $B$).  We see that the training objective
decomposes to a trace-norm regularized problem in $A$ and a trace-norm
regularized problem in $B$.  Each one of these problems is a
trace-norm regularized learning problem, under a uniform sampling
distribution (in the corresponding submatrix) of a noisy low-rank
``orthogonal'' matrix, and can therefor be learned with $\tO(k n_A)$
and $\tO(k n_B)$ samples respectively.  In other words, $\tO(k n)$
samples should be enough to learn both inside $A$ and inside $B$.

However, the regularization tradeoff parameter $\lambda$ compounds the
two problems.  When the objective is expressed in terms of
$\tc(\cdot)$, as in \eqref{eq:trdecomp1}, the regularization tradeoff
is scaled differently in each part of the training objective.  With
$\tO(k n)$ samples, it is possible to learn in $A$ with some setting
of $\lambda$, and it is possible to learn in $B$ with some other
setting of $\lambda$, but from the discussion above we learn that no
single value of $\lambda$ will allow learning in both $A$ and $B$.
Either $\lambda$ is too high yielding too strict regularization in
$B$, so learning on $B$ is not possible, perhaps since it is scaled by
$n_B\gg n_A$.  Or $\lambda$ is too small and does not provide enough
regularization in $A$.

\begin{figure}[t!]
\vspace{0.0in}
\hbox{ \centering \hspace{-0.1in}
\setlength{\epsfxsize}{3.1in}
\epsfbox{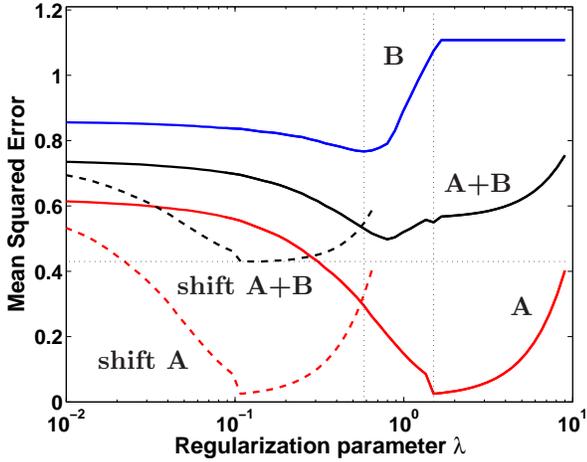}
 \hspace{0.2in}
\begin{picture}(0,0)(0,0)
\put(-100, +150){\textrm{{\bf \normalsize B}}}
\put(-77, +102){\textrm{{\bf \normalsize A+B}}}
\put(-52, +55){\textrm{{\bf \normalsize A}}}
\put(-208, +35){\textrm{{\bf \normalsize shift A}}}
\put(-178, +63){\textrm{{\bf \normalsize shift A+B}}}
\end{picture}
}
\vspace{-0.05in} 
\caption{ \label{fig:sim} \small
The solid curves show the optimum of the mean squared error
objective \eqref{eq:trdecomp1} (unweighted trace-norm),
as a function of the regularization parameter $\lambda$.
The dashed curves display a weighted trace-norm.
}
\vspace{-0.0in}
\end{figure}

Returning to our simulation experiment, 
the solid curves of Fig. \ref{fig:sim}
show the excess test error for the
minimizer of the training objective \eqref{eq:trdecomp1}, as a
function of the regularization tradeoff parameter $\lambda$.  Note
that these are essentially the same curves as displayed in Fig. \ref{fig:tc},
except the path of regularized solutions is now parameterized by
$\lambda$ rather then by the bound on $\tc(X)$.  Not surprisingly we
see the same phenomena: different values of $\lambda$ are required for
optimal learning on $A$ and on $B$.  Forcing the same $\lambda$ on
both parts of the training objective \eqref{eq:trdecomp1} yields
a deterioration in the generalization performance.

\section{Weighted Trace Norm}\label{sec:weighted}
The decomposition \eqref{eq:trdecomp1} and the discussion in the
previous section suggests weighting the trace-norm by the frequency of
rows and columns.  For a sampling distribution $\dist$, denote by
$p(i)$ the row marginal, i.e.~the probability of observing row $i$,
and similarly denote by $q(j)$ the column marginal.  We propose using the
weighted version of the trace-norm as a regularizer:
\beqa \label{eq:ww} 
  \trpqnorm{X}{p,q} = \trnorm{\diag(\sqrt{p}) X
    \diag(\sqrt{q})}\notag \hspace{0.6in} \\
=\min_{X=U'V} \frac{1}{2}(\sum_i p(i) \norm{U_i}^2 + \sum_j q(j), 
\norm{V_j}^2)
\eeqa 
where $\diag(\sqrt{p})$ is a diagonal matrix with $\sqrt{p(i)}$ on its
diagonal (similarly $\diag(\sqrt{q})$).  The corresponding normalized
complexity measure is given by $\tc_{p,q}(X) = \trpqnorm{X}{p,q}^2$.
Note that for a uniform distribution we have that
$\tc_{p,q}(X)=\tc(X)$.  Furthermore, it is easy to verify that for an
``orthogonal'' rank-$k$ matrix $X$ we have $\tc_{p,q}(X) = k$ for {\em
  any} sampling distribution.

Equipped with the weighted trace-norm as a regularizer, let us revisit
the problematic sampling distribution studied in the previous Section.
In order to fit the ``orthogonal'' rank-k $X^*$, we need a weighted
trace-norm of $\trpqnorm{X^*}{p,q} = \sqrt{\tc_{p,q}(X)} = \sqrt{k}$.
How large a sample $S \cap A$ can we now shatter using such a weighted
trace-norm?  We can shatter a sample if $\trnorm{Y_{S \cap A}} \leq
\sqrt{k}$.  In order to calculate $\trnorm{Y_{S \cap A}}$, recall that
for $(i,j)\in A$ we have $p(i)=q(j)=1/(2 n_A)$.  We can now calculate:
$\trpqnorm{Y_{S \cap A}}{p,q} = \trnorm{\sqrt{1/(2n_A)} Y_{S \cap A}
  \sqrt{1/(2n_A)}} = \trnorm{Y_{S \cap A}} / (2n_A) \leq \sqrt{|S \cap A|
  n_A}/(2n_A) = \sqrt{|S|/(8n_A)}$.  That is, we can
shatter a sample of size up to $|S| = 8 k n_A < 8 k n$.  The calculation
for $B$ is identical.  It seems that now, with a fixed constraint on the
weighted trace-norm, we have enough capacity to both fit $X^*$, and
with $\tO(k n)$ samples, avoid overfitting on $A$.

Returning to the penalization view \eqref{eq:cC} we can again
decompose the training objective:
\begin{equation}\label{eq:wtrlamba1}
  \min_X \frnorm{Y_S-X_S}^2 + \lambda \trpqnorm{X}{p,q}, 
\end{equation}
\vspace{-0.01in} 
as: 
\vspace{-0.01in} 
\begin{align}
& \frnorm{Y_S-X_S}^2 + \lambda \trpqnorm{X}{p,q} \notag\\
& = (\frnorm{Y_{A\cap S}-X_{A\cap S}}^2 + \lambda \trpqnorm{X_A}{p,q})
\notag\\
&\quad +
  (\frnorm{Y_{B\cap S}-X_{B \cap S}}^2 + \lambda \trpqnorm{X_B}{p,q})\notag \\
& = \left(\frnorm{Y_{A\cap S}-X_{A\cap S}}^2 + \lambda/2
\sqrt{\tc_A(X_A)}\right) \notag\\
&\quad+
  \left(\frnorm{Y_{B\cap S}-X_{B \cap S}}^2 + \lambda/2 \sqrt{\tc_B(X_B)}\right)
  \label{eq:wtrdecomp1}
\end{align}
avoiding the scaling by the block sizes which we encountered in
\eqref{eq:trdecomp1}.

Returning to the synthetic experiments of Fig. \ref{fig:sim}, 
and comparing \eqref{eq:trdecomp1} with
\eqref{eq:wtrdecomp1}, we see that introducing the weighting
corresponds to a relative change of $n_A/n_B$ in the correspondence of
the regularization tradeoff parameters used for $A$ and for $B$. This
corresponds to a shift of $\log\frac{n_A}{n_B}$ in the log-domain used
in the figure.  Shifting the solid red (bottom) curve by this amount
yields the dashed red (bottom) curve.  The solid blue (top) curve and
the dashed red (bottom) curve thus represent the excess error on $B$
and on $A$ when the weighted trace norm is used, i.e. the training
objective \eqref{eq:wtrdecomp1} is minimized (except for an overall
scaling in $\lambda$).  The dashed black (middle) curve is the overall
excess error when using this training objective.  As can be seen, the
weighting aligns the excess errors on $A$ and on $B$ much better, and
yields a lower overall error. The weighted trace-norm achieves
the lowest MSE of 0.4301 with corresponding $\lambda = 0.11$.  This is
compared to the lowest MSE of 0.4981 with $\lambda=0.80$, achieved by
the unweighted trace-norm.  It is also interesting to observe that the
weighted trace-norm outperforms its unweighted counterpart for a wide
range of regularization parameters $\lambda \in [0.01; 0.6]$. This may
also suggest that in practice, particularly when working with large
and imbalanced datasets, it may be easier to
search for regularization parameters using weighted trace-norm.
Fig. \ref{fig:tc}, right panel, further shows the test error
as a function on the constraint $\tc_{p,q}(X)$.

Finally, Fig. \ref{fig:sim} also suggests that the optimal shift is
actually smaller then $n_A/n_B$.  We consider a smaller shift
by using the partially-weighted trace-norm:
\beqa
  \trpqnorm{X}{p,q,\alpha} = \trnorm{\diag(p^{\alpha/2}) X
    \diag(q^{\alpha/2})}\notag \hspace{0.4in} \\
=\min_{X=U^{\top}V} \frac{1}{2}(\sum_i p(i)^{\alpha} \norm{U_i}^2 + \sum_j q(j)^{\alpha}
\norm{V_j}^2)
\eeqa 
And he corresponding normalized
complexity measure $\tc_{p,q,\alpha}(X) =
\trpqnorm{X}{\frac{p^{\alpha}}{n^{1-\alpha}},\frac{q^{\alpha}}{m^{1-\alpha}}}^2$.

\section{Practical Implementation}
When dealing with large datasets, such as the Netflix data, the most
practical way to fit trace-norm regularized models is through
stochastic gradient descent \cite{SalMnih08,KorenSVDpp}.

Let $n_i = \sum_j S_{ij}$  and $m_j = \sum_i S_{ij}$
denote the number of observed ratings for user $i$ and movie $j$ respectively. 
The training objective (over the index pairs $(i,j)$) 
using partially-weighted trace-norm (Eq. \ref{eq:ww}) can be written as: 
\beqa \label{eq:sgd} 
  \sum_{\{i,j\}\in S}  \bigg( \big(Y_{ij} - U_i^{\top} V_j\big)^2 + \hspace{1.2in}  \\ 
  +\frac{\lambda}{2} \bigg(\frac{ p(i)^\alpha}{n_i}    \norm{U_i}^2
+\frac{q(j)^\alpha}{m_j}  \norm{V_j}^2\bigg) \bigg),  \nonumber
\eeqa
where $U \in \R^{k\times n}$ and $V \in \R^{k \times m}$.  
We can
optimize this objective using stochastic gradient descent by picking
one training pair $(i,j)$ at random at each iteration, and taking a step in
the direction opposite the gradient of the term corresponding to the
chosen $(i,j)$.

Note that even though the objective \eqref{eq:sgd} as a function of
$U$ and $V$ is non-convex, there are no non-global local minima if we
set $k$ to be large enough, i.e.~$k > \min(n,m)$
\cite{burer2005local}.  However, fitting orthogonal models in practice
with very large values of $k$ becomes computationally expensive.
Instead, we consider truncated trace-norm minimization by restricting
$k$ to smaller values. In the next section we demonstrate that even
when using truncated trace-norm, its weighted version significantly
improves model's prediction performance.

In all of our experiments, we also replace unknown row $p(i)$ and
column $q(j)$ marginals in $\eqref{eq:sgd}$ by their empirical estimates
$\hat{p}(i) = \nicefrac{n_i}{|S|}$ and $\hat{q}(j) =
\nicefrac{m_j}{|S|}$. This results in the following objective: \beqa
\label{eq:sgd2}
\sum_{\{i,j\}\in S}  \bigg( \big(Y_{ij} - U_i^{\top} V_j\big)^2 + \hspace{1.2in}  \\
+\frac{\lambda}{2|S|} \bigg( n_i^{\alpha-1} \norm{U_i}^2
+m_j^{\alpha-1} \norm{V_j}^2\bigg) \bigg).  \nonumber \eeqa Setting
$\alpha=1$, corresponding to the weighted trace-norm \eqref{eq:ww},
results in stochastic gradient updates that do not involve the row and
column counts at all and are in some sense the simplest.  Strangely,
and likely originating as a ``bug'' in calculating the stochastic
gradients by one of the participants, these are the actual SGD steps
used by many practitioners on the Netflix dataset
\cite{KorenSVDpp,GravityJMLR,SalMnih08}.


\section{Experimental results}
We evaluated various models on the Netflix dataset, which is the largest publicly
available collaborative filtering dataset.
The training set contains
100,480,507 ratings from 480,189 randomly-chosen, anonymous users
on 17,770 movie titles. As part of the training data, Netflix also
provides qualification set, containing 1,408,395 ratings.
The pairs were selected from the most recent ratings for a subset of the
users in the training dataset. 
Due to the special selection scheme, ratings from users with few ratings are
overrepresented in the qualification set, relative to the training set. 
To avoid the issue of dealing with different training and
test distributions, 
we also created our own validation and test sets, each containing
100,000 ratings that were 
randomly selected from the training set.  
As a baseline, Netflix provided the test score of its own system
trained on the same data, which is 0.9514.

This dataset is interesting for several reasons. First, it is very large, and very
sparse (98.8\% sparse).  Second, the dataset is very imbalanced with
highly non-uniform samples.
It includes users with over 10,000 ratings as well as users who rated fewer than
5 movies. 

\subsection{Results} 
In our first experiment, 
for various values of $\alpha$,
we fit parameters $U$ and $V$ using stochastic gradient descent as in 
\eqref{eq:sgd2} with $k=30$.   
Both $U$ and $V$  were randomly initialized 
for all models and regularization parameters $\lambda$ were chosen by cross-validation.

\begin{table}[t]
\vspace{-0.05in} 
{\small
\caption{ \label{tab:res} \small
Model performance  using 
Root Mean Squared Error (RMSE)  
on the Netflix qualification set and the test set, that was 
randomly subsampled from the training data.
}
\label{tab:results}
\begin{center}
\hspace{-0.2in} 
\begin{tabular}{c|ccc|ccc}
\hline 
   &           &   \multicolumn{2}{c}{RMSE}  & & \multicolumn{2}{c}{RMSE} \\
$\alpha$ & k   &  Test & Qual  & k &  Test  & Qual \\ \hline
1    &  30  & 0.7607  & 0.9105  & 100 &  0.7412 & 0.9071 \\
0.9  &  30  & 0.7573  & 0.9091  & 100 &  0.7389 & 0.9062 \\
0.75 &  30  & 0.7723  & 0.9128  & 100 &  0.7491 & 0.9098 \\
0.5  &  30  & 0.7823  & 0.9159  & 100 &  0.7613 & 0.9127 \\
0    &  30  & 0.7889  & 0.9235  & 100 &  0.7667 & 0.9203 \\ \hline
\end{tabular}
\end{center}
}
\vspace{-0.15in} 
\end{table}

Performance results of the weighted trace-norm regularization
for various values of $\alpha$ are
shown in table~\ref{tab:res}.
Observe that that the weighted trace-norm ($\alpha=1$) 
achieved a RMSE of 0.9105 on the Netflix qualification set, 
significantly outperforming its unweighted counterpart with $\alpha=0$,
that achieved a RMSE of 0.9235. This large performance gap is
striking. It clearly suggests that the weighting is quite important.
Table \ref{tab:res} further reveals that the 
weighted trace-norm ($\alpha=1$) is not
optimal. Surprisingly, partially weighted trace-norm with $\alpha=0.9$
 achieved a RMSE of 0.9091, slightly outperforming the weighted matrix factorization.
Performance results on the artificially created test set 
are similar to the results on the qualification set. 
Note also that the large gap in generalization performance between
the test and the qualification sets is due to 
the Netflix's special qualification selection scheme. 

In our second experiment, we fitted much larger models with $k=100$. 
As expected, the weighted trace-norm regularization ($\alpha=1$) 
attained a RMSE 0.9071, 
significantly improving upon the unweighted model's RMSE of 0.9203. 
Again, this large performance gap strongly suggests that the 
weighting can yield significant performance boost, particularly when dealing with very 
imbalanced data, such as the Netflix dataset. 

In all of our experiments, 
we also empirically observed that for a wide range of regularization
parameters $\lambda$, optimizing the weighted trace-norm 
almost always yielded better predictions on both the test and the Netflix qualification sets
than optimizing the unweighted trace-norm.
This confirms our previous results on the synthetic experiment
and strongly suggests that it may be far easier to
search for regularization parameters using the weighted trace-norm.

\section{Discussion}
In this paper we showed both analytically and empirically that 
under non-uniform sampling, trace-norm regularization 
can lead to significant performance deterioration and an increase
in sample complexity. Motivated by our analytic analysis, 
we further suggested a corrected version of the 
trace-norm, called weighted trace-norm, that does take into account the
non-uniform sampling distribution.  
Our results on both synthetic and highly imbalanced 
Netflix datasets further demonstrate that 
the weighted trace-norm yields significant improvements in prediction quality.  
It is interesting to note that setting $\alpha=1$ in the weighted
trace-norm objective \eqref{eq:ww} implies that the frequent users
(movies) get regularized much stronger than the rare users (movies).
From Bayesian perspective, such regularization is quite unusual, since
it effectively states that the effect of the prior becomes stronger as
we observe more data.  Yet, our analysis and empirical results
strongly suggest that in non-uniform setting, such ``unorthodox''
regularization is crucial for achieving good generalization
performance.

Although theoretical guarantees are not the focus of this work, we
hope that the weighted trace-norm, and the discussions in Sections
\ref{sec:fails} and \ref{sec:weighted}, will be helpful in deriving
theoretical learning guarantees for non-uniform sampling
distributions, both in the form of generalization error bounds as in
\cite{srebro2005rank}, and generalizing the compressed-sensing
inspired work on recovery of noisy low-rank matrices as in 
\cite{candes2009matrix,recht-preprint}.

\subsection*{Acknowledgments}
R.S. acknowledges the financial support from NSERC, Shell, and NTT
Communication Sciences Laboratory.

 
{ 
\bibliography{tc}
\bibliographystyle{icml2010}
}
\end{document}